\documentclass{article} 
\usepackage[preprint]{colm2026_conference}

\usepackage[T1]{fontenc}
\usepackage[utf8]{inputenc}
\usepackage{microtype}
\usepackage{hyperref}
\usepackage{url}
\usepackage{booktabs}
\usepackage{amsmath,amssymb,amsfonts}
\usepackage{graphicx}
\usepackage{subcaption}
\usepackage{xcolor}
\usepackage{lineno}

\graphicspath{{figures/}}

\definecolor{darkblue}{rgb}{0, 0, 0.5}
\hypersetup{colorlinks=true, citecolor=darkblue, linkcolor=darkblue, urlcolor=darkblue}

\title{Reward-Informed Sparse Autoencoders and the Solution-Completeness Confound}

\author{%
\makebox[0.5\textwidth][c]{%
\begin{tabular}{c}
Tanvi Nagilla$^{*}$ \\
\texttt{nagillatanvi@gmail.com} \\[1.2ex]
Daniel Manta \\
\texttt{26dmanta@fmschools.org}
\end{tabular}}%
\makebox[0.5\textwidth][c]{%
\begin{tabular}{c}
Alexander Jameson$^{*,\dagger}$ \\
\texttt{alex@phanguard.org} \\[1.2ex]
Shayaan Uddin \\
\texttt{shayuddin4532@gmail.com}
\end{tabular}}%
}

\begin{document}

\ifcolmsubmission
\fi

\maketitle

\renewcommand{\thefootnote}{\fnsymbol{footnote}}
\setcounter{footnote}{0}
\footnotetext[1]{Lead authors; equal contribution.}
\footnotetext[2]{Corresponding author.}
\renewcommand{\thefootnote}{\arabic{footnote}}
\setcounter{footnote}{0}

\begin{abstract}
Sparse autoencoders (SAEs) decompose language-model activations into sparse, interpretable features, and an appealing way to aim them at reasoning is to curate their data with a signal reinforcement learning already produces: the reward. We build such a \emph{reward-informed} SAE (RI-SAE): we split GRPO trajectories into high-reward (``good'') and low-reward (``bad'') reasoning continuations, train a standard JumpReLU SAE on their activations, and then ask what the resulting good/bad separation actually measures. On Llama-3.1-8B a sparse subset of the $16{,}384$ features does separate the classes (silhouette $0.79$ on the selected features versus $0.005$ for the full code), but a control battery shows the separation is largely \emph{solution completeness} rather than reasoning quality: a TF-IDF text classifier already splits the classes (AUC $0.75$--$0.83$), and three structural cues alone (length, a closed reasoning block, and a boxed answer) reach AUC $0.70$ ($99\%$ of good versus $69\%$ of bad completions are boxed). A generic SAE that never saw the reward does not separate the classes at all (silhouette $0.01$, no discriminative features), so the $0.79$ is in-sample fitting of this curated signal rather than structure that a reward-blind dictionary recovers. We therefore present the recipe and its control battery together: reward filtering is a cheap, label-free way to reuse RL signals for interpretability, but most of what it surfaces is completion form. Two discriminative features are still readable (symbolic mathematics; procedural and evaluative language), which we take as illustrative rather than as isolated reasoning.
\end{abstract}

\section{Introduction}

Language models solve multi-step reasoning problems, but the internal computations behind a correct derivation are hard to read off from activations \citep{wei2023chainofthoughtpromptingelicitsreasoning,geiping2025scalingtesttimecomputelatent}. Sparse autoencoders address this by reconstructing a layer's activations under a sparsity penalty, recovering dictionaries of features that are often monosemantic \citep{bricken2023monosemanticity,lieberum2024gemmascope}. But SAEs are typically trained on undifferentiated text: they learn whatever reconstructs the activation distribution, with no preference for the features that separate competent from incompetent reasoning.

A signal that already encodes that distinction sits unused in RL pipelines: the reward. Reinforcement learning with verifiable rewards, and GRPO in particular, produces many model rollouts, each tagged with a scalar reward \citep{shao2024deepseekmath}. We treat this reward as a cheap form of supervision for interpretability and use it to decide which activations an SAE should look at.

We study the simplest version of this idea, a \emph{reward-informed} SAE (RI-SAE). The reward enters through the data, not the objective: we keep high-reward GRPO continuations as ``good reasoning'' and low-reward ones as ``bad reasoning'', and train an otherwise standard SAE on their activations. No new loss, no retraining of the base model; the method can be attached to an existing RL run. The recipe works in the narrow sense that its features separate the classes; the harder question, and our main one, is what that separation actually measures. Our contributions are: (i) a control battery that answers it, showing the good/bad split is largely \emph{solution completeness} (whether a complete, well-formed answer was produced) rather than reasoning quality, with a TF-IDF text baseline at AUC $0.75$--$0.83$, structure-only features at $0.70$, and a generic reward-blind SAE that does not separate the classes at all (so the effect is in-sample fitting, not a generic artifact); (ii) the reward-filtering recipe itself, a cheap, label-free way to turn an RL reward into SAE supervision; and (iii) two interpretable features and a Gemma-2-2B GSM8K fine-tuning study that sketch where the method could go. We see RI-SAEs less as a finished diagnostic than as a reusable signal whose results must be read against the control battery we provide.

\section{Method}
\label{sec:method}

\paragraph{Reward-based data curation.} We build the corpus from a public pool of GRPO continuations, each with a scalar reward in $[-1, 3]$ and a \texttt{<think>}$\ldots$\texttt{</think>}\texttt{<answer>}$\ldots$\texttt{</answer>} format. The raw pool contains substantial non-English (largely Chinese) text, which we remove with a non-ASCII filter. We then label \textbf{good reasoning} ($1$) as reward $\geq 2.0$ and \textbf{bad reasoning} ($0$) as reward $\leq 0.5$; both require at least $30$ tokens and must pass a coherence check (no excessive $n$-gram repetition or canned refusals), so the bad class is genuinely low-reward reasoning rather than broken text. Our main set is balanced at $1{,}000$ good and $1{,}000$ bad (mean reward $2.83$ vs.\ $0.06$); good completions are longer than bad (median $1{,}211$ vs.\ $453$ words). Two caveats are built in: the labels are a reward proxy (``non-reasoning'' is shorthand for low-reward reasoning), and the reward enters only here, in data selection.

\paragraph{Sparse autoencoder.} For a residual-stream activation $x\in\mathbb{R}^d$, the SAE computes $z=\mathrm{ReLU}(W_{\text{enc}}x+b_{\text{enc}}-\theta)$ and $\hat{x}=W_{\text{dec}}z+b_{\text{dec}}$, where $\theta$ is a learned per-feature threshold (JumpReLU) and $W_{\text{dec}}$ has unit-norm rows. We train with the standard objective
\begin{equation}
\mathcal{L} = \lVert x-\hat{x}\rVert_2^2 + \lambda\lVert z\rVert_0, \qquad \lambda=0.1,
\label{eq:loss}
\end{equation}
using Adam (lr $3\times10^{-4}$, batch $16$) for $500$ steps. The reward does not appear in Eq.~\ref{eq:loss}. Feature analyses use a wide, overcomplete $16{,}384$-feature dictionary (roughly $4\times$ expansion for Llama-3.1-8B), the regime in which interpretable features have been reported \citep{lieberum2024gemmascope,rajamanoharan2024jumprelu}; activations are taken from layer~$22$. A reward-weighted variant of Eq.~\ref{eq:loss} is a straightforward extension we have not implemented.

\section{Results}

\subsection{What does the good/bad separation measure?}

\begin{figure}[t]
\centering
\begin{subfigure}[b]{0.46\linewidth}
\includegraphics[width=\linewidth]{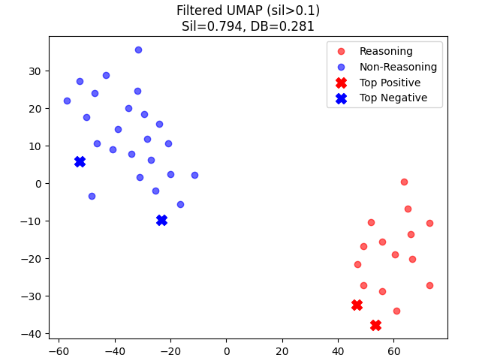}
\caption{}
\label{fig:umap}
\end{subfigure}\hfill
\begin{subfigure}[b]{0.46\linewidth}
\includegraphics[width=\linewidth]{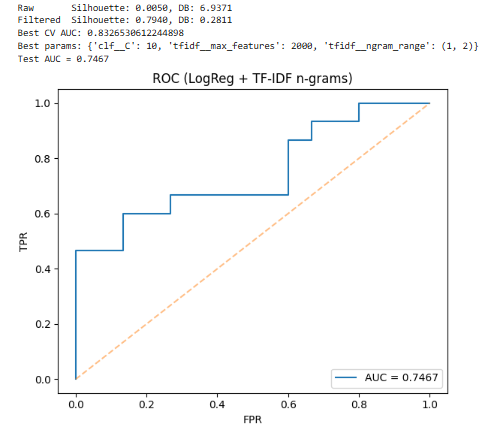}
\caption{}
\label{fig:logreg}
\end{subfigure}
\caption{Separating good- from bad-reasoning trajectories (Llama-3.1-8B, layer~22). (a) UMAP of the SAE features that individually discriminate the classes; the selected subspace clusters cleanly (silhouette $0.79$), while the full code gives $0.005$. (b) A TF-IDF $n$-gram classifier on the raw text reaches AUC $0.83$ (CV) / $0.75$ (held out), showing a strong lexical confound.}
\label{fig:separation}
\end{figure}

Reward-filtered features do separate the two classes, but only after selection and only in a way a plain text classifier matches. In the full $16{,}384$-dimensional code the classes do not separate (silhouette $0.005$, Davies--Bouldin $6.94$). Keeping the features that individually discriminate them (per-feature silhouette $>0.1$) and embedding that subspace with UMAP yields clean clusters (silhouette $0.79$, Davies--Bouldin $0.28$; Figure~\ref{fig:separation}a). That jump is produced by the selection step (we pick discriminative features and then report separation on them), so it shows a \emph{sparse subset} of features carries the distinction, not that the SAE separates reasoning globally.

\paragraph{The separation is mostly solution completeness.} That distinction is mostly structural. High- and low-reward completions differ in vocabulary, formatting, and length, not only in reasoning, and a control battery (Table~\ref{tab:confound}) quantifies it. A TF-IDF $n$-gram classifier on the raw text reaches AUC $0.83$ cross-validated and $0.75$ held out (Figure~\ref{fig:separation}b); removing digits or answer-formatting tokens barely moves it (it stays near $0.83$--$0.85$); and three structural features alone (length, whether the reasoning block is closed, and whether a \verb|\boxed{}| answer appears) reach AUC $0.70$. The cues are concrete: the top good-class $n$-grams are \texttt{boxed}, \texttt{final answer}, and \texttt{therefore}; $99\%$ of good versus $69\%$ of bad completions contain a boxed answer; and $100\%$ versus $83\%$ close the reasoning block. Much of ``good versus bad reasoning'' is thus \emph{solution completeness}, which bounds how much of any reward-filtered SAE result can be read as reasoning rather than completion form.

\paragraph{A generic SAE does not recover the separation.} Is the discriminative subspace produced by reward filtering, or by the SAE alone? We ran the same pipeline on a generic, pretrained Llama Scope SAE for the matched Llama-3.1-8B residual stream \citep{he2024llamascope}, one trained on ordinary text that never saw the reward. On the same $1{,}000/1{,}000$ set it does not separate the classes: full-code silhouette $0.012$ (cf.\ $0.005$), \emph{no} feature exceeds our per-feature selection threshold (max single-feature silhouette $0.059$), total-activation AUC $0.61$, and selecting and UMAP-embedding its $50$ most class-different features still yields silhouette $0.02$. The $0.79$ is therefore neither intrinsic to these activations nor a generic UMAP artifact; it requires an SAE trained on the curated set itself. Because that training and the feature selection are in-sample, and the distinction is largely completeness, we read the $0.79$ as in-sample fitting of a completeness-dominated signal rather than reward-independent reasoning structure. This control isolates the SAE from the data but not reward filtering from in-domain training; a same-recipe SAE on \emph{unfiltered} in-domain data would isolate it, and is left to future work.

\begin{table}[t]
\centering
\caption{Anatomy of the good/bad separation (Llama-3.1-8B; cross-validated AUC of a logistic classifier on the named features). Stripping lexical content barely changes the separation, and three structural cues alone (length, a closed block, a boxed answer) recover most of it.}
\label{tab:confound}
\begin{tabular}{lc}
\toprule
Features & AUC \\
\midrule
Full text (TF-IDF $n$-grams)                          & $0.84 \pm 0.03$ \\
\quad digits removed                                  & $0.85 \pm 0.02$ \\
\quad answer/structure cues removed                   & $0.83 \pm 0.03$ \\
\quad digits \emph{and} cues removed                  & $0.83 \pm 0.02$ \\
Structure only (length, closed block, boxed answer)   & $0.70 \pm 0.20$ \\
\quad length only                                     & $0.60$ \\
\bottomrule
\end{tabular}
\end{table}

\subsection{Two discriminative features are interpretable}

\begin{figure}[t]
\centering
\begin{subfigure}[b]{0.48\linewidth}
\includegraphics[width=\linewidth]{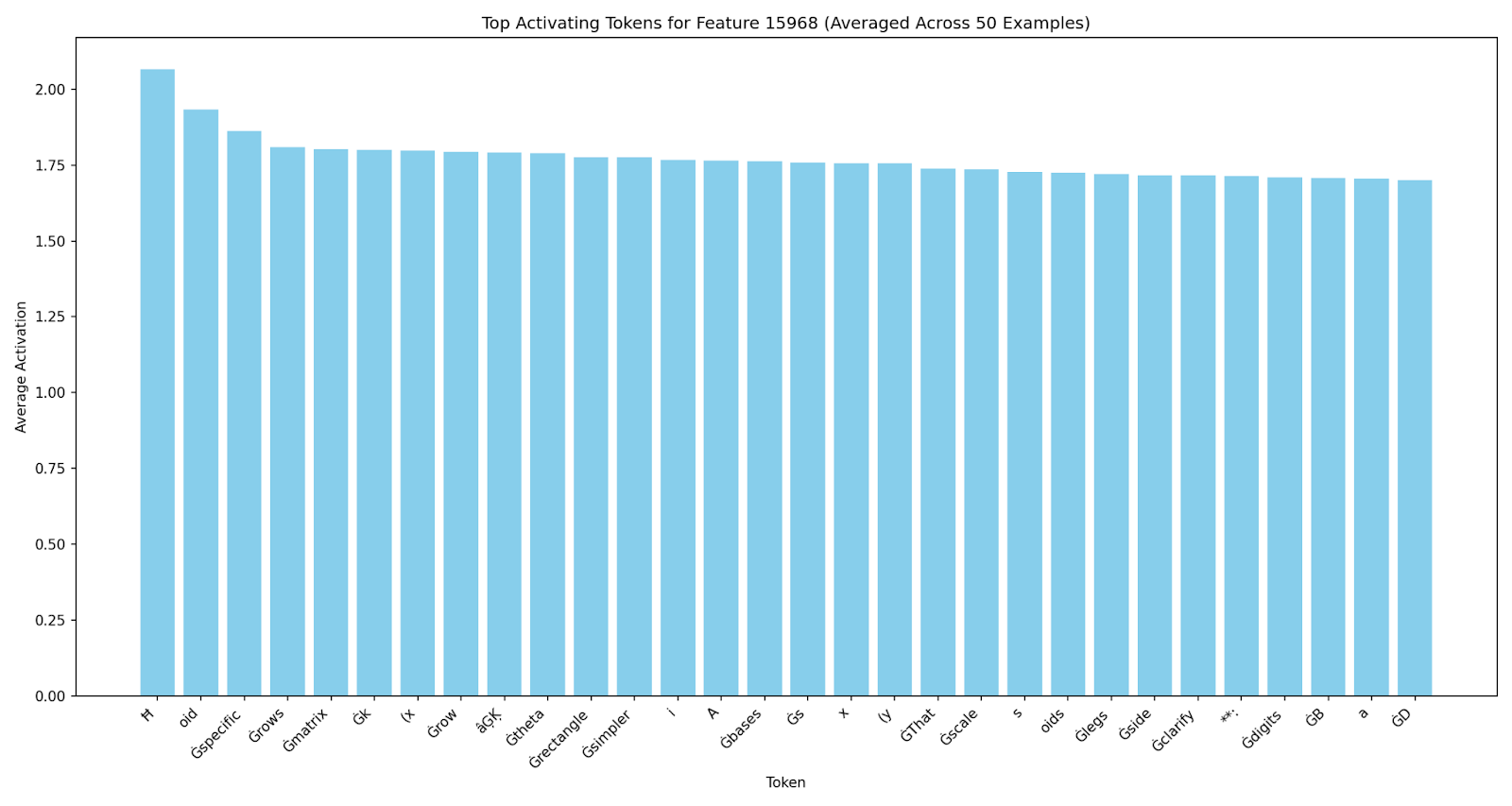}
\caption{}
\label{fig:f15968}
\end{subfigure}\hfill
\begin{subfigure}[b]{0.48\linewidth}
\includegraphics[width=\linewidth]{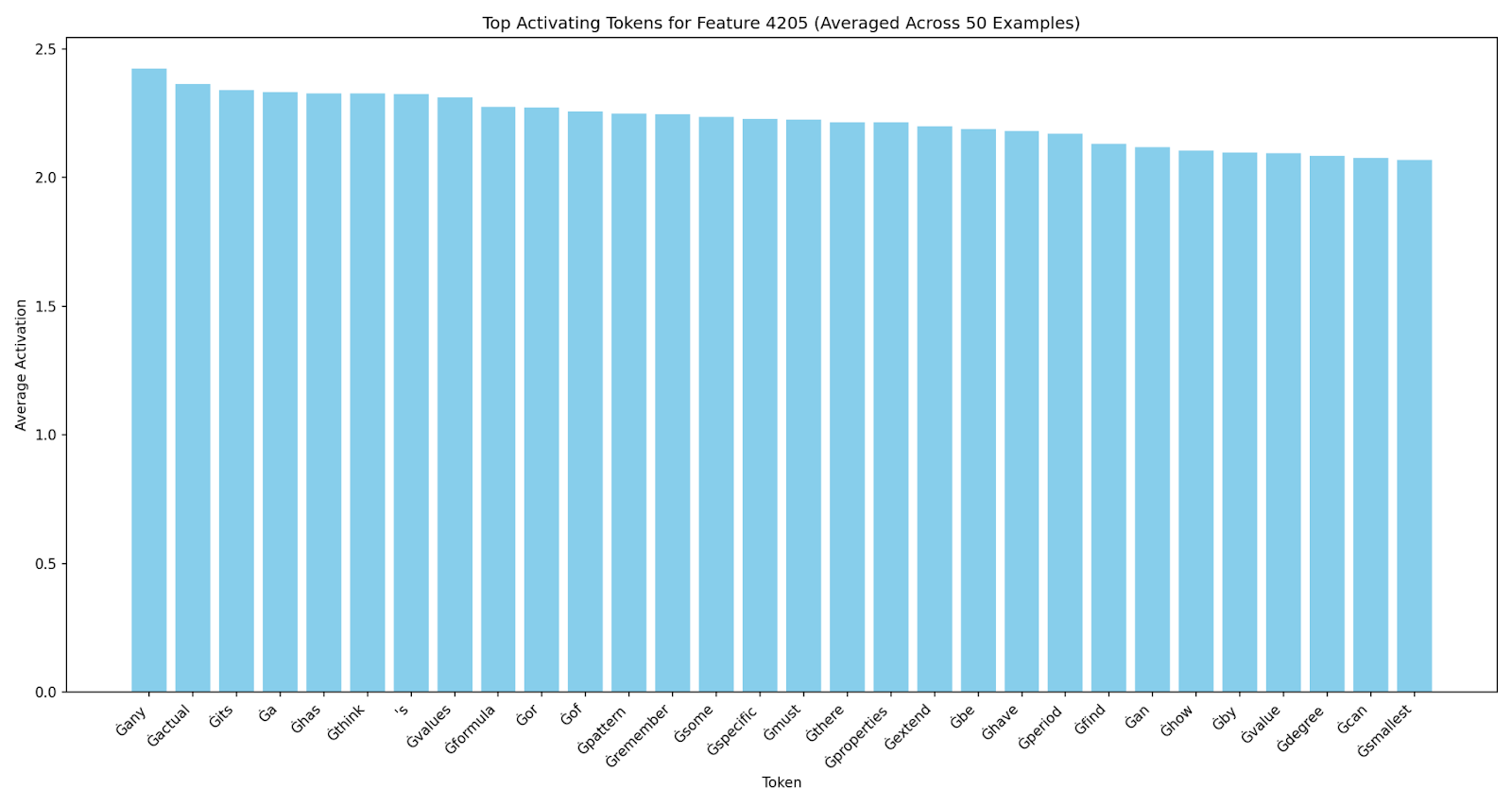}
\caption{}
\label{fig:f4205}
\end{subfigure}
\caption{Top-activating tokens for two discriminative features (Llama-3.1-8B, layer~22, averaged over $50$ examples). (a) Feature $15968$: symbolic mathematics. (b) Feature $4205$: procedural/evaluative language (with some common function words).}
\label{fig:features}
\end{figure}

Ranking features by the difference in mean activation between the classes gives consistent but modest differences ($\approx\pm0.1$), spread across many features. We interpret two highly discriminative ones by their top-activating tokens, computed by running the encoder over the tokens of $50$ examples and averaging each feature's strongest tokens (Figure~\ref{fig:features}; in the byte-level BPE tokenizer a leading space is written as a special prefix symbol). Feature $15968$ fires on symbolic-mathematics tokens (\texttt{matrix}, \texttt{rows}, \texttt{theta}, \texttt{(x}, \texttt{digits}), a clean example of a structured-mathematics feature. Feature $4205$ is associated with procedural and evaluative language (\texttt{think}, \texttt{values}, \texttt{formula}, \texttt{remember}, \texttt{must}), though it also picks up common function words, so its interpretation is suggestive rather than definitive. Starting only from a reward signal, we arrive at named features a practitioner can inspect; we do not claim they are causally responsible for reasoning.

\subsection{Toward monitoring fine-tuning}

The use we ultimately have in mind is monitoring reasoning features as a model trains. As a first step we fine-tuned Gemma-2-2B on GSM8K, saved five checkpoints (steps $600$--$3000$), and evaluated each (Appendix~\ref{app:finetune}). Exact match rises from $0.5\%$ to $\sim$5\% and MAE on numeric answers falls from $\sim$4950 to $\sim$100, with most of the gain between steps $1200$ and $1800$. This characterizes the model, not its features: we did not run an SAE across checkpoints. It does, however, give a ready setup: training an RI-SAE on these checkpoints to test whether reasoning features sharpen in that same window is the immediate next step.

\section{Limitations}

The labels are a reward proxy, not reasoning versus its absence. The solution-completeness confound (Table~\ref{tab:confound}) is the paper's main result rather than a caveat, and it bounds the rest: on these numbers alone we cannot read the SAE separation as reasoning structure. The headline silhouette is computed on features selected for being discriminative and so reflects that selection, not global separation ($0.005$ on the full code). Evaluation sets are small and results come from single runs without seeds or error bars, so we present them as descriptive. Our generic-SAE control shows the SAE alone does not drive the separation, but it does not isolate reward filtering from in-domain training; the cleanest remaining test (a same-recipe SAE on \emph{unfiltered} in-domain data), along with causal interventions and a concept-alignment metric \citep{fel2025archetypal}, is future work. The reward-weighted objective is so far only a proposal, and the feature analyses (Llama-3.1-8B) and fine-tuning study (Gemma-2-2B) use different backbones.

\section{Conclusion}

Reward signals are an underused resource for interpretability, but they are not a free one. Filtering RL trajectories by reward and training a standard SAE does surface a sparse set of features that separate good from bad reasoning on Llama-3.1-8B, yet our control battery shows that most of that separation is solution completeness, recoverable from length, a closed reasoning block, and a boxed answer (AUC $0.70$) and already matched by a plain text classifier. The honest reading is that reward filtering is a cheap, reusable way to point an SAE at reasoning-adjacent data, but its good/bad signal must be read against a completeness baseline before any feature is called a reasoning feature; the battery we report is the tool for doing so. The natural next steps sharpen the test rather than the claim: a same-recipe SAE on unfiltered in-domain data to isolate filtering from in-domain training (a generic SAE already fails to separate the classes); a reward-weighted objective; and SAE-based monitoring across fine-tuning checkpoints.

\subsubsection*{Acknowledgments}

We thank Kevin Zhu and Ryan Lagasse for their guidance and feedback throughout this project.

\bibliography{references}
\bibliographystyle{colm2026_conference}

\appendix

\section{Fine-tuning details and curves}
\label{app:finetune}

We fine-tuned Gemma-2-2B on GSM8K with LoRA and evaluated five checkpoints on the GSM8K test set, scoring exact match (EM) on the final answer and mean absolute error (MAE) on the numeric answer. EM is low in absolute terms because of a strict last-line string match; MAE shows the numeric answers nonetheless converge toward the correct values.

\begin{figure}[h]
\centering
\begin{subfigure}[b]{0.42\linewidth}
\includegraphics[width=\linewidth]{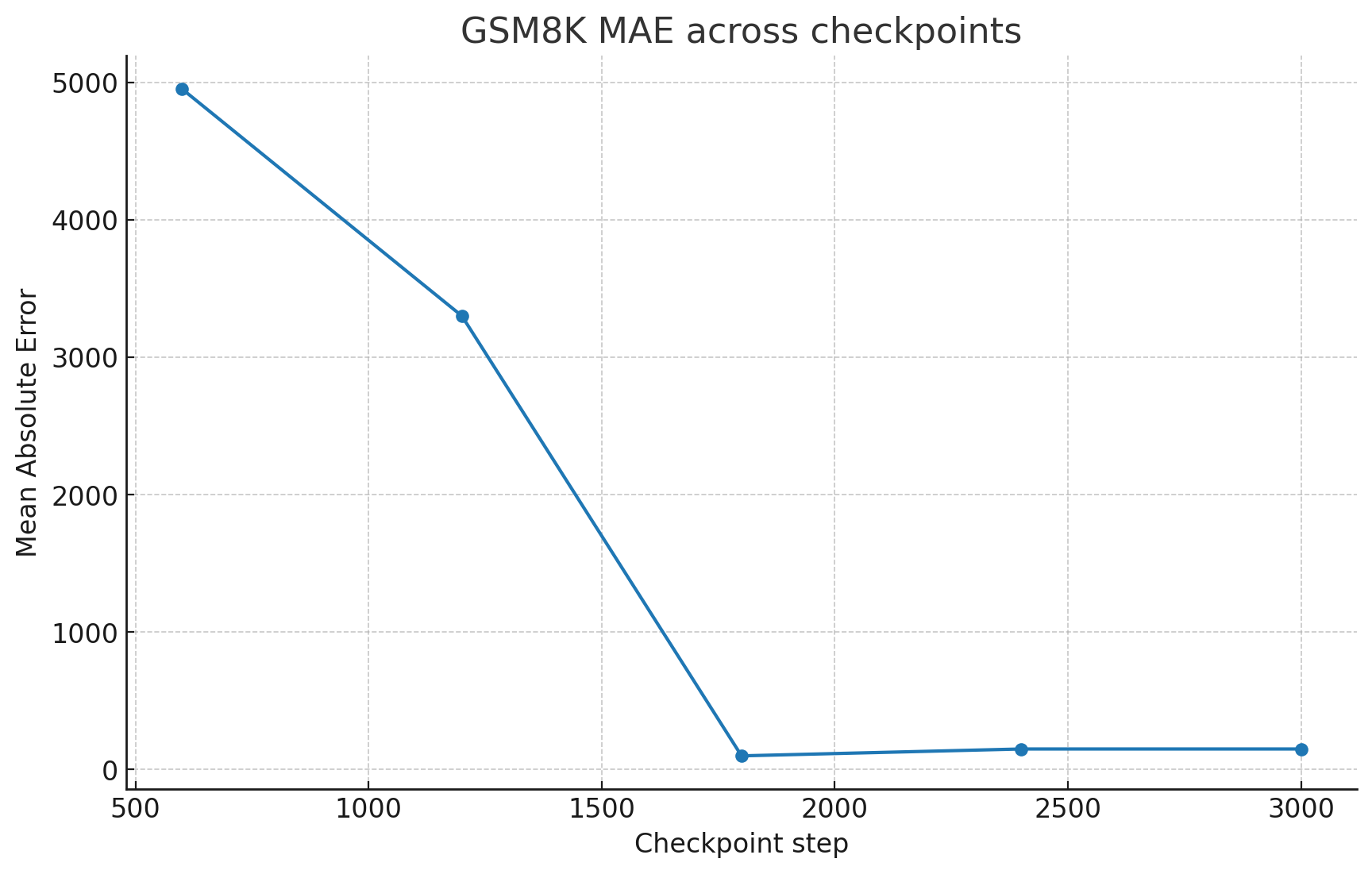}
\caption{}
\end{subfigure}\hfill
\begin{subfigure}[b]{0.42\linewidth}
\includegraphics[width=\linewidth]{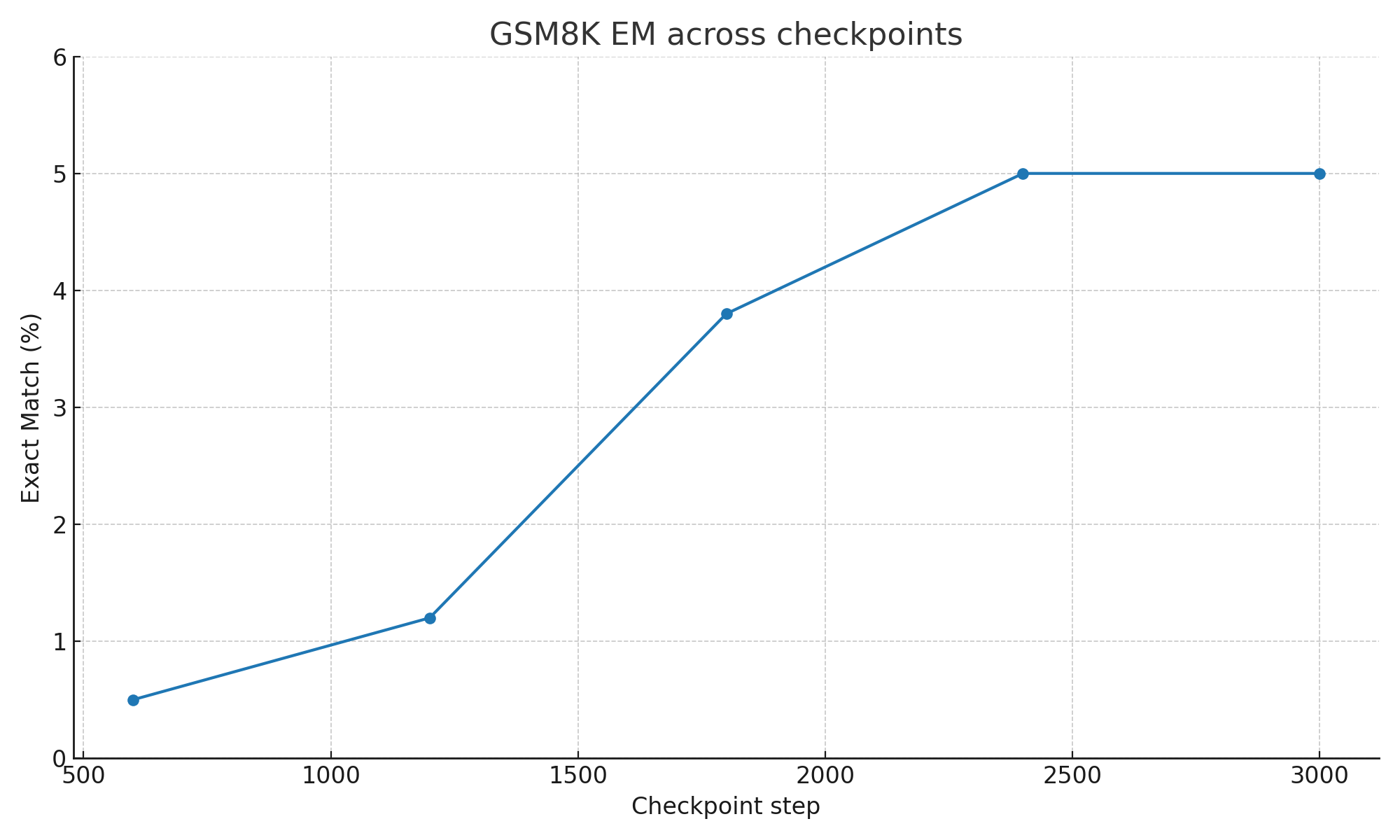}
\caption{}
\end{subfigure}

\vspace{2mm}

\begin{subfigure}[b]{0.7\linewidth}
\includegraphics[width=\linewidth]{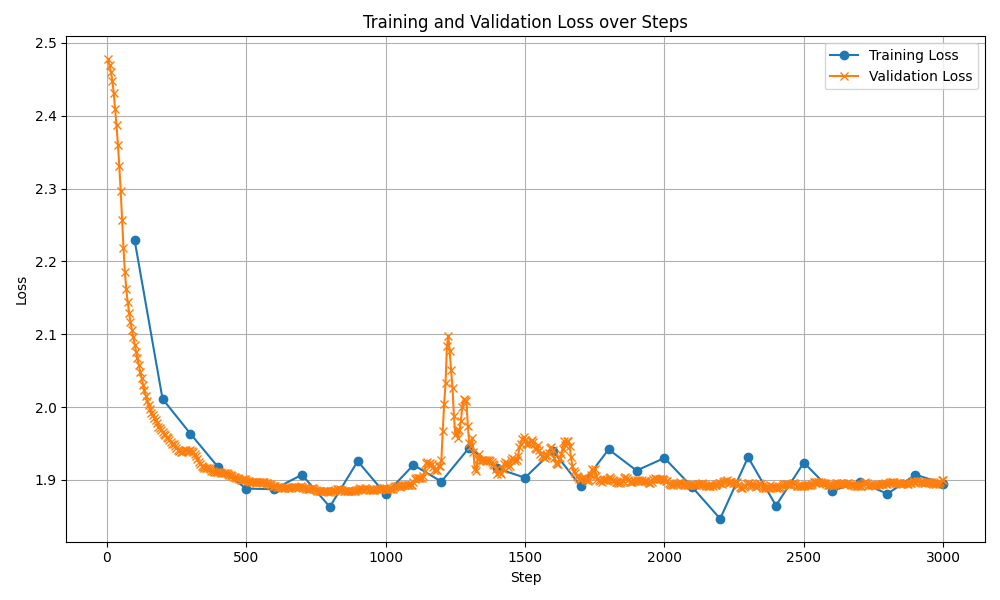}
\caption{}
\end{subfigure}
\caption{Gemma-2-2B GSM8K fine-tuning. (a) MAE on numeric answers. (b) Exact match. (c) Training/validation loss. These are model-performance curves; SAE-based feature tracking across the checkpoints is left to future work.}
\label{fig:finetune}
\end{figure}

\section{SAE configuration}

JumpReLU SAE (learned per-feature threshold initialized at $0.001$, unit-norm decoder rows); objective MSE $+\,0.1\lVert z\rVert_0$; Adam, lr $3\times10^{-4}$, batch $16$, $500$ steps; dictionary width $16{,}384$; Llama-3.1-8B activations at layer~22 for the reported analyses. Data curation as in Section~\ref{sec:method}: reward thresholds $2.0$ / $0.5$, minimum $30$ tokens, trigram-repetition and refusal-phrase coherence filters.

\end{document}